%% file: main.tex
\documentclass[10pt,conference,letterpaper]{IEEEtran}
\IEEEoverridecommandlockouts
\usepackage{amsmath,amssymb,amsfonts}
\usepackage{graphicx}
\usepackage{textcomp}
\usepackage{xcolor}
\def\BibTeX{{\rm B\kern-.05em{\sc i\kern-.025em b}\kern-.08em T\kern-.1667em\lower.7ex\hbox{E}\kern-.125emX}}
\usepackage{svg}
\usepackage[T1]{fontenc} % Use 8-bit encoding that has 256 glyphs
\usepackage{microtype} % Slightly tweak font spacing for aesthetics

\usepackage[english]{babel} % Language hyphenation and typographical rulest has 256 glyphs
\usepackage{microtype} % Slightly tweak font spacing for aesthetics

\usepackage[english]{babel} % Language hyphenation and typographical rules

\usepackage{caption}
\usepackage{subcaption}
\usepackage{siunitx}
\usepackage{gensymb}
\usepackage{booktabs}
\usepackage{adjustbox}
\usepackage{tikz}
\usepackage{svg}
\usepackage{soul}
\graphicspath{ {images/} }

\usepackage[font=small,labelfont=bf]{caption}

\usepackage{glossaries}
\glsdisablehyper
\loadglsentries[main]{acronyms}
\begin{document}

% \title{Learning Congestion Control in Tactical Scenarios: \\ A Reinforcement Learning Environment with Container Networks}
%\title{[WIP] Learning to Sail Dynamic Networks:\\ A Reinforcement Learning \hl{Agent} Framework for Congestion Control in Tactical Environments}
\title{Learning to Sail Dynamic Networks:\\ The MARLIN Reinforcement Learning Framework for Congestion Control in Tactical Environments}

% ----------------- AUTHORS HERE ------------
\author{

\IEEEauthorblockN{
Raffaele Galliera\IEEEauthorrefmark{1}\IEEEauthorrefmark{2},
Mattia Zaccarini\IEEEauthorrefmark{3},
Alessandro Morelli\IEEEauthorrefmark{1},
Roberto Fronteddu\IEEEauthorrefmark{1}, \\
Filippo Poltronieri\IEEEauthorrefmark{3},
Niranjan Suri\IEEEauthorrefmark{1}\IEEEauthorrefmark{4},
Mauro Tortonesi\IEEEauthorrefmark{3}
}

\IEEEauthorblockA{\IEEEauthorrefmark{1} 
Florida Institute for Human \& Machine Cognition (IHMC)\\
Email: \{rgalliera, amorelli, rfronteddu, nsuri\}@ihmc.org}

\IEEEauthorblockA{\IEEEauthorrefmark{2}
Department of Intelligent Systems \& Robotics - The University of West Florida (UWF), Pensacola, FL, USA \\
Email: rg101@students.uwf.edu}

\IEEEauthorblockA{\IEEEauthorrefmark{3}
Distributed Systems Research Group, University of Ferrara, Ferrara, Italy \\
Email: \{mattia.zaccarini, filippo.poltronieri, mauro.tortonesi\}@unife.it}

\IEEEauthorblockA{\IEEEauthorrefmark{4}
US Army DEVCOM Army Research Laboratory (ARL), Adelphi, MD, USA \\
Email: niranjan.suri.civ@army.mil}
} % end -- author block

\maketitle

\begin{abstract}
% Tactical networks pose distinct challenges for Congestion Control (CC) compared to traditional networks. However, conventional algorithms, such as TCP Cubic, struggle in these environments due to limited resources, frequent disconnections, and variable connectivity. Recent efforts have explored the use of Reinforcement Learning (RL) for CC, but they often struggle to generalize, particularly in competitive, unstable, and unforeseen scenarios. To the best of our knowledge, none of the proposed approaches have focused on their deployment in tactical networks and the typical challenges entailed by such dynamic environments. This paper builds upon MARLIN, an RL-based CC agent designed for real networks, to explore its potential in dynamic environments like tactical networks. To adapt MARLIN to the specific needs of tactical networks, we modify its RL formulation and incorporate dynamic link conditions that resemble the challenging nature of these scenarios. We evaluate our agent in file transfer tasks and compare its performance against TCP Cubic and other tactical communication middleware. The results highlight the effectiveness of our agent in optimizing CC in unreliable tactical networks. This work represents a promising step toward leveraging RL to address their unique challenges, bridging the gap between RL and complex networking scenarios.
Conventional \gls{cc} algorithms, such as TCP Cubic, struggle in tactical environments as they misinterpret packet loss and fluctuating network performance as congestion symptoms. 
Recent efforts, including our own MARLIN, have explored the use of \gls{rl} for \gls{cc}, but they often fall short of generalization, particularly in competitive, unstable, and unforeseen scenarios. 
To address these challenges, this paper proposes an \gls{rl} framework that leverages an accurate and parallelizable emulation environment to reenact the conditions of a tactical network. We also introduce refined \gls{rl} formulation and performance evaluation methods tailored for agents operating in such intricate scenarios.
We evaluate our \gls{rl} learning framework by training a MARLIN agent in conditions replicating a bottleneck link transition between a \gls{satcom} and an \gls{uhf} radio link. Finally, we compared its performance in file transfer tasks against \gls{tcp} Cubic and the default strategy implemented in the Mockets tactical communication middleware.
The results demonstrate that the MARLIN \gls{rl} agent  outperforms both \gls{tcp} and Mockets under different perspectives and highlight the effectiveness of specialized \gls{rl} solutions in optimizing \gls{cc} for tactical network environments. 
\end{abstract}


\section{Introduction}
    \label{sec:intro}
    \input{sections/introduction}
\section{Congestion Control in Tactical Networks}
    \label{sec:scenario}
    \input{sections/scenario}

\section{Leveraging Container Networks to Emulate Challenging Networking Scenarios}
    \label{sec:emulated_marlin}
    \input{sections/emulated_marlin}
    
\section{A Reinforcement Learning Formulation for Congestion Control in Tactical Environments}
    \label{sec:juice}
    \input{sections/juice}

\begin{figure}[t] % <---
\centering
   \begin{subfigure}{0.35\textwidth}
   \centering
       \includegraphics[width=\columnwidth]{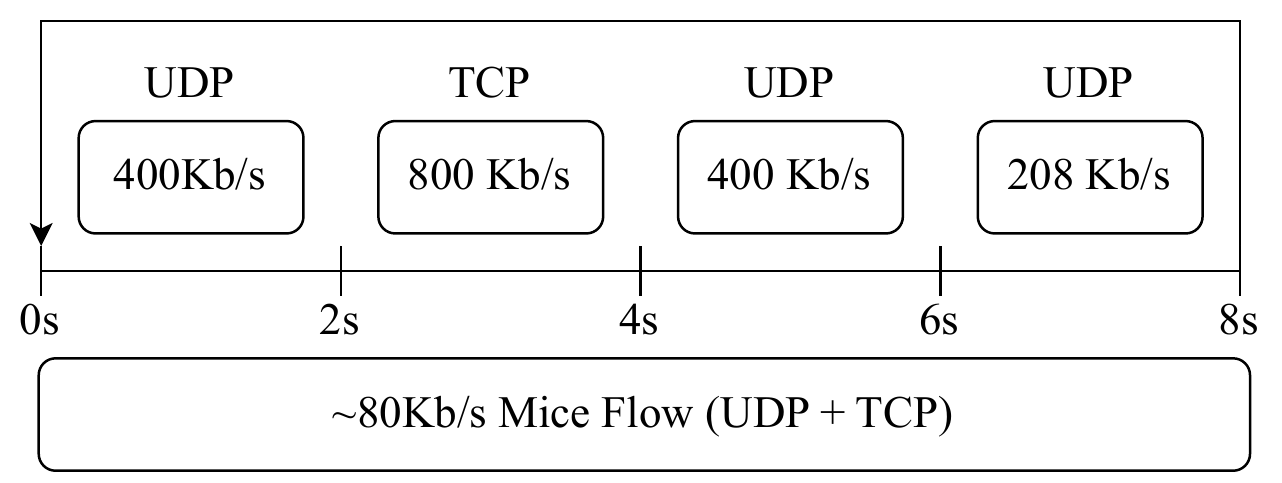}
       \caption{Background traffic during the \gls{satcom} link usage}
       \label{fig:satcom_traffic}
   \end{subfigure}
\hfill % <--- 
   \begin{subfigure}{0.35\textwidth}
   \centering
       \includegraphics[width=\columnwidth]{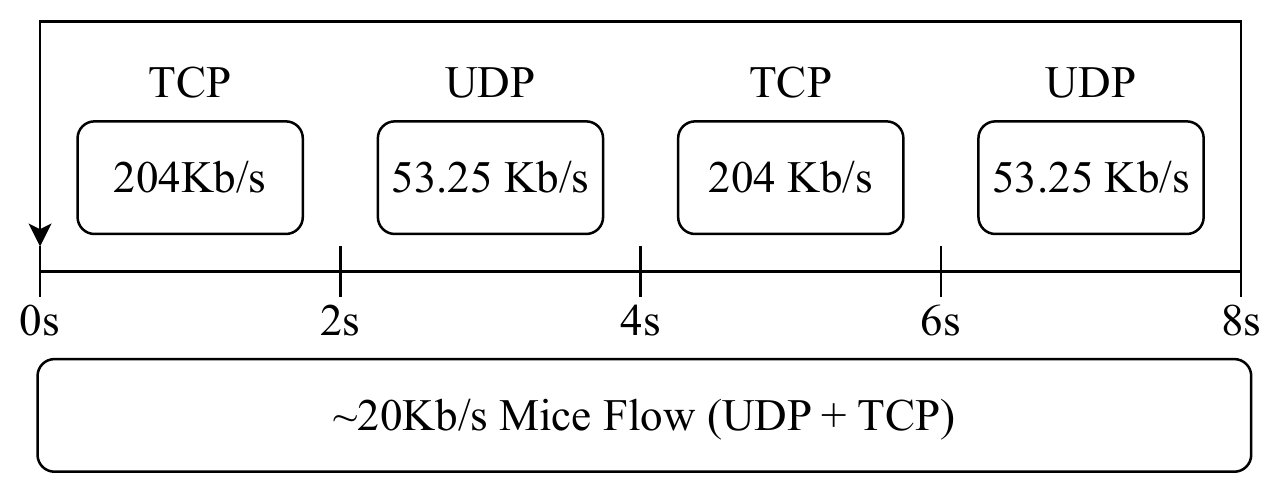}
       \caption{Background traffic during the \gls{uhf} link usage}
       \label{fig:uhf_traffic}
   \end{subfigure}
   \caption{Characterization of the background traffic patterns.}
   \label{fig:traffic_patterns}
\end{figure}

\section{Experimental Results}
    \label{sec:results}
    \input{sections/experimental_results}

\section{Related Work}
    \input{sections/related_work}

\section{Conclusion and Future Work}
    \input{sections/conclusion}

\bibliographystyle{IEEEtran}
\bibliography{bibliography}

%\newpage
%\printbibliography

\end{document}

%% file: acronyms.tex
\newacronym{c5isr}{C5ISR}{Command, Control, Communications, Computers, Cyber, Intelligence, Surveillance, and Reconnaissance}
\newacronym{cwnd}{CWND}{Congestion Window}
\newacronym{rl}{RL}{Reinforcement Learning}
\newacronym{rtt}{RTT}{Round-Trip Time}
\newacronym{srtt}{SRTT}{Smoothed Round-Trip Time}
\newacronym{bdp}{BDP}{Bandwidth Delay Product}
\newacronym{mdp}{MDP}{Markov Decision Process}
\newacronym{sb3}{SB3}{Stable-Baselines3}
\newacronym{ack}{ACK}{ACKnowledge}
\newacronym{rpc}{RPC}{Remote Procedure Call}
\newacronym{sac}{SAC}{Soft Actor-Critic}
\newacronym{ema}{EMA}{Exponential Moving Average}
\newacronym{ml}{ML}{Machine Learning}
\newacronym{ai}{AI}{Artificial Intelligence}
\newacronym{cc}{CC}{Congestion Control}
\newacronym{tcp}{TCP}{Transmission Control Protocol}
\newacronym{udp}{UDP}{User Datagram Protocol}
\newacronym{rl3}{RL-Zoo3}{RL Baselines3 Zoo}
\newacronym{mgen}{MGEN}{Multi-Generator Network Test Tool}
\newacronym{ap}{AP}{Access Point}
\newacronym{cots}{COS}{Commercial Off-The-Shelf}
\newacronym{rpi}{RPi4B}{Raspberry Pi 4 Model B}
\newacronym{qos}{QoS}{Quality Of Service}
\newacronym{peb}{PEB}{Partial Episode Bootstrapping}
\newacronym{scp}{SCP}{Secure Copy Protocol}
\newacronym{netem}{TC-NETEM}{TC NetEm - Traffic Control Network Emulator}
\newacronym{asic}{ASIC}{Application-Specific Integrated Circuit}
\newacronym{sgd}{SGD}{Stochastic Gradient Descent}
\newacronym{marl}{MARL}{Multi-Agent Reinforcement Learning}
\newacronym{nic}{NIC}{Network Interface Card}
\newacronym{abc}{ABC}{Appropriate Byte Counting}
\newacronym{ss}{SS}{Slow Start}
\newacronym{satcom}{SATCOM}{Satellite Communication}
\newacronym{uhf}{UHF}{UHF Wide Band}
\newacronym{uav}{UAV}{Unmanned Aerial Vehicle}
\newacronym{rti}{RTI}{RTT Transition Impact}
\newacronym{notm}{NOTM}{Networking On The Move}
\newacronym{hq}{HQ}{headquarters}
\newacronym{dl}{DL}{Deep Learning}

%% file: sections/introduction.tex
Tactical networks present unique challenges for \gls{cc}. While operating in demanding circumstances marked by 
limited resources, unreliable links, frequent disconnections, and varying levels of connectivity,
these networks need to support critical real-time functionalities to facilitate mission applications such as \gls{c5isr}. 

Traditional \gls{cc} algorithms, such as those embedded in transport protocols like \gls{tcp}, face significant difficulties in maintaining efficient communications within tactical networks, as they were developed for wired environments and misinterpret as congestion symptoms phenomena such as packet losses and temporary unreachability, that are very common in tactical networks, thus severely and unnecessarily reducing transmission speeds. Optimizing \gls{cc} in unreliable networks necessitates innovative approaches that are able to cope with their dynamic and unpredictable nature.

% RL is a promising approach
In this context, the combination of \gls{rl} techniques with \gls{dl} for policy parameterization, often referred to as Deep Reinforcement Learning, has emerged as a promising approach. Deep \gls{rl} demonstrated remarkable robustness across diverse domains, and its application in computer networks offers new possibilities for addressing the challenges of \gls{cc}~\cite{Sivakumar2019,drl-cc}. By leveraging \gls{rl}, agents can be trained to learn optimal policies through interactions with the network environment, enabling more efficient and reliable communications.

Despite numerous efforts showing promising results, the performance of these \gls{rl} agents still falls short of generalization capabilities, especially when unreliable and unpredictable links are encountered. This performance gap can be attributed to various factors either linked to the learning problem itself, such as designing the environment where the agent resides and its "learning curriculum", or to the challenging and partially observable nature of the networking environment. Addressing these challenges requires the development of solutions that enable \gls{rl} agents to learn effective policies for \gls{cc} decision-making while experiencing complex and dynamic scenarios. However, creating dedicated environments for conducting experiments in such scenarios can be challenging, costly, and even infeasible in real network environments. To overcome this last constraint, the importance of an accurate emulation environment cannot be overstated.

This paper extends our work within the MARLIN project ~\cite{galliera2023marlin} by proposing an \gls{rl} framework that leverages an accurate and parallelizable emulation environment to reenact the conditions of a tactical network, thus allowing agents to experience a wide range of dynamic behaviors. 
To better evaluate the decision-making effectiveness of our agent, we also introduce a novel metric, the \gls{rti}, based on the maximum \gls{rtt} detected during a communication involving a link transition, that allows to measure the agent's responsiveness to link changes and its queue management capabilities.

% The primary focus of this work is to extend the training framework introduced in \cite{galliera2023marlin} by incorporating dynamic link conditions into the training scenarios. 
% This approach would constitute a first natural step following the lessons learned in real-world networks. 

To evaluate our learning framework, we trained an \gls{rl} agent in an emulated environment replicating a bottleneck link transition between a \gls{satcom} link and an \gls{uhf} radio link - rather typical in tactical networks. 
We then evaluated the agent performance in a file transfer task and compared it against conventional \gls{cc} algorithms such as \gls{tcp} Cubic~\cite{cubic}, as well as \gls{cc} algorithms implemented within communication middlewares tailored for tactical environments such as Mockets~\cite{mockets1}, which the same MARLIN utilizes as partnering protocol. 
The results demonstrate that the exposure of the agent to complex networking scenarios enables training policies able to achieve competitive decision-making performance, and validate the \gls{rl} training approach based on accurate emulation and purposely designed valuation metrics for specialized environments like tactical networks.

%% file: sections/scenario.tex
Tactical networks represent a unique communications environment due to a combination of particularly harsh and dynamic network conditions paired with a high degree of heterogeneity in the network technologies employed and, consequently, the characteristics of the links. This situation demands efficient data transmission strategies that can quickly respond to changes in the status of the network, which makes the role of transport protocols and \gls{cc} algorithms of crucial importance in tactical networks and especially challenging.

Traditional and de-facto standard \gls{cc} algorithms for \gls{tcp}, such as CUBIC, were not designed for this type of scenario and, as a consequence, tend to underperform significantly or, in some cases, cause connections to break altogether, leading to significant loss of network resources \cite{mockets2}. Frequent packet loss causes loss-based \gls{cc} algorithms, which interpret loss as a sign of congestion, to step out of the slow start phase and decrease their congestion window, impeding the effectiveness of the bandwidth discovery process. In contrast, delay-based \gls{cc} algorithms typically misinterpret abrupt changes in the end-to-end latency, such as those caused by changes in the routing rules, as a sign that they are allowed to either increase or reduce their \gls{cwnd} significantly. This behavior can again lead to severe underutilization of already scarce network resources.

This fostered the design of dedicated solutions to overcome the limitations of \gls{tcp} in tactical environments, such as Mockets \cite{mockets1}, a \gls{udp}-based communication middleware that allows extensive customization of its configuration parameters. By enabling adjustments to parameters like the \gls{cwnd}, the Pending Packet Queue Size, and the Selective Acknowledgment Transmission Timeout, Mockets offers the flexibility to fine-tune its settings according to the specific requirements of different communication channels.
Mockets implements a very aggressive \gls{cc} algorithm whose purpose is to fully utilize the link bandwidth in the presence of variable communication latency and elevated packet loss. While it presents several advantages over \gls{tcp} and other transport protocols in degraded environments~\cite{mockets2}, the existing \gls{cc} fails to share link capacity with other communication flows going through the same links and cannot adapt quickly to changes in the available bandwidth.

% scenario
These limitations are especially visible in case of heterogeneous network technologies. % such as \gls{satcom} and \gls{uhf} radios. 
In fact, abrupt changes from high-bandwidth, high-latency links like \gls{satcom} to a low-bandwidth, low-latency links like \gls{uhf}, and back, can wreak havoc on the performance of \gls{cc} solutions. 
%\hl{lead to long-lasting queue build-ups (bufferbloat) and congestion in the reversed scenario.}
% The highly dynamic behaviour of these environments may lead to situations where the \gls{satcom} link becomes unavailable or its performance degrades abruptly.
To compound these issues, tactical environments exhibit: routing rules that can change during a mission;
% (e.g., in the case of a \gls{satcom} reachback link that becomes unusable due to worsening weather conditions, some or all network communications might be rerouted over a lower capacity, lower latency \gls{uhf} radio link)
highly varying wireless performance; node mobility; and the enemy's activity, which could jam communications and/or render nodes unreachable.
%inoperational 
These challenging scenarios necessitate the utilization of advanced techniques capable of enabling dynamic and responsive reconfiguration in response to evolving network conditions.

% The characteristics that render tactical networks difficult environments for traditional transport protocols and \gls{cc} algorithms, such as \gls{tcp} CUBIC, include very limited bandwidth, high and variable latency (jitter), unpredictable packet loss, and elevated churn rate. 

%% file: sections/emulated_marlin.tex
\begin{figure}
    \centering   
    \includegraphics[width=\columnwidth] {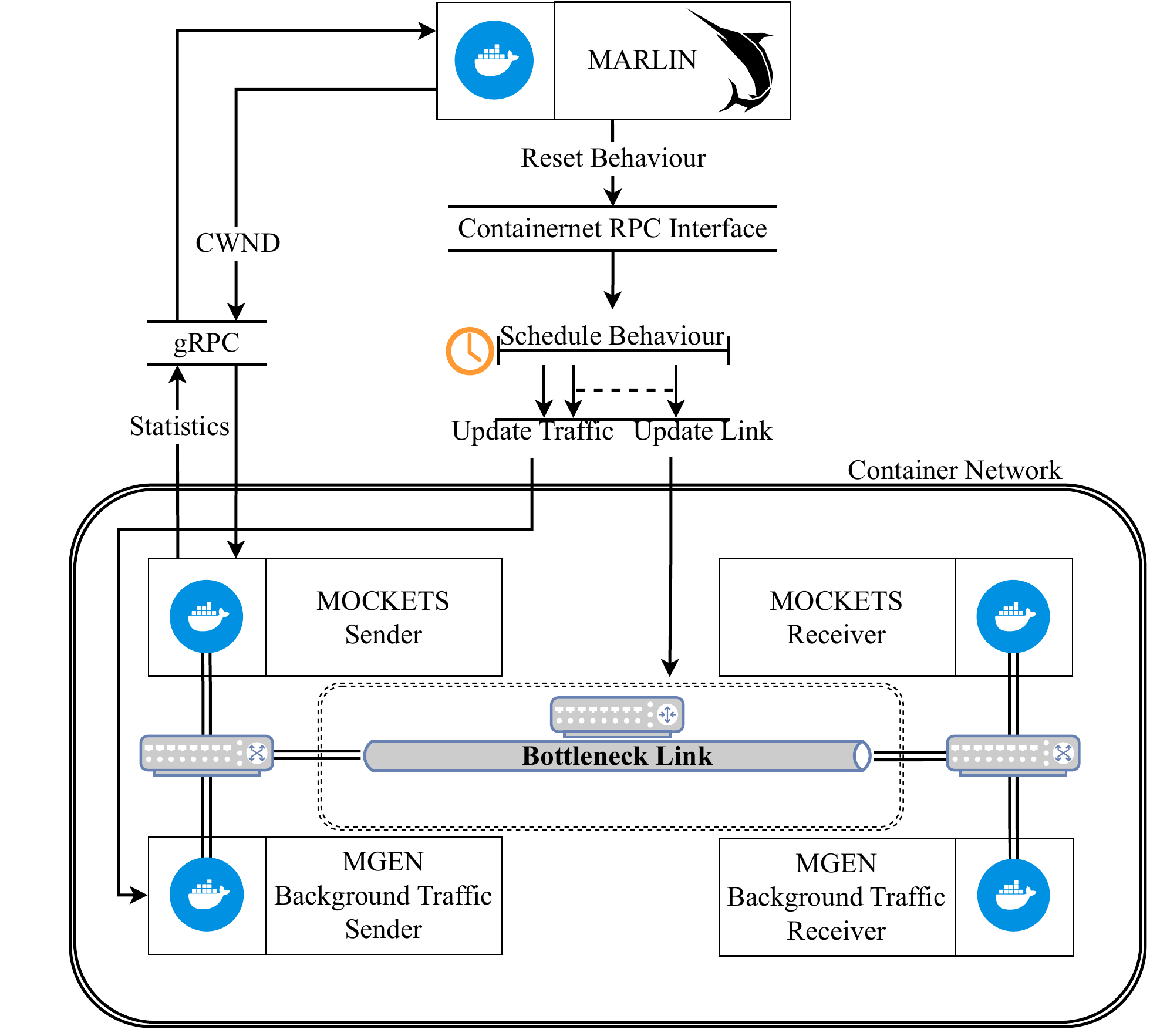}
    \caption{The MARLIN containerized training framework. 
    %container-based architecture
    }
     \label{fig:architecture}
\end{figure}
% such asincluding limited control, variability, reproducibility, and scalability issues

% When such challenges merge with the unique, constrained, and complex networking scenarios faced in tactical networks, reproducing such interactions for training purposes becomes unfeasible. 

% Expand the motivation of your approach
Complex networking scenarios, such as those encountered in tactical environments, are known for their intricacy, often involving sets of heterogeneous entities with distinct behaviors~\cite{anglova}. However, \gls{rl} environments dedicated to \gls{cc} are usually not designed to capture such complexities in their training setting, making the agents prone to underperform when deployed in these scenarios. 

To bridge this gap, our work enhances the MARLIN framework\footnote{Code available at https://github.com/RaffaeleGalliera/marlin-rlcc} by incorporating customizable link behaviors, which in this case we implement to resemble the complexities of tactical environments. By introducing these diverse scenarios into the training environment, we can expose the \gls{rl} agent to a wider range of experiences, enabling it to learn and adapt to the intricate behaviors exhibited by the heterogeneous entities in tactical networks. 

Specifically, we extend the MARLIN solution to employ a network generator module. In succinct terms, such a component is responsible to create custom container networks, define the behavior of the links involved, and generate background traffic from competitive/collaborative sources. The reader can refer to Figure~\ref{fig:architecture} for a summary of the framework, which also includes the topology defined for this work (detailed in \ref{sec:results}).
    
The network generator module defines a \gls{rpc} interface exposing a series of methods bonded to a Containernet~\cite{peuster2016medicine} instance operating on the host machine. Containernet is a lightweight container-based network emulation tool built on Mininet that allows users to create and manage virtual networks with Docker containers for the purpose of simulating complex network scenarios.

Once started, our network generator module defines a starting topology running specific Docker images for every container involved. The \gls{rpc} methods are then used by the \gls{rl} environment to create and define diverse scenarios. Every time the environment is initialized or reset, a dedicated \gls{rpc} method is invoked, setting the links to their original values. In the meantime, the same environment will start the communication between the Mockets sender and its receiving counterpart, which effectively starts the collaboration between the agent and the protocol. Other hosts are prompted to generate traffic, competing with the \gls{rl} agent for link usage. These containers will induce different \gls{tcp} and \gls{udp} traffic behaviors following a \gls{mgen}~\cite{mgen} script defined through the \gls{rpc} arguments.
    
In Section~\ref{subsec:reward}, we detail how such a controlled setting might guide the agent toward effective policies.

%% file: sections/juice.tex
In this section, we present the fundamental \gls{rl}~\cite{suttonBarto} components on which this version of MARLIN relies. In particular, we address the challenge of determining the timing of actions in the tactical scenario and propose a reward function specifically designed for these complex environments.

% \subsection{Congestion Control as a Markov Decision Process}
% \label{subsubsec:ccasrl}
%     Transport protocols like \gls{tcp} implement traditional \gls{cc} algorithms that adjust their \gls{cwnd} based on environmental changes to optimize throughput and minimize congestion. These changes can include packet loss, duplicate packets, or variations in \gls{rtt}. This sequential decision-making process can be formalized using the \gls{mdp}~\cite{puterman1994} framework, which encompasses the agent-environment interaction through the tuple $\langle S, A, p, R \rangle$. At each time step $t$, three signals are exchanged: the agent's action $a_t \in A(s_t)$, based on the observed state $s_t \in S$, results in a reward signal $r_{t+1} \in R$ and a transition to a new state $s_{t+1}$. The fundamental principles of learning from interaction in \gls{rl}~\cite{suttonBarto} stem from the MDP abstraction, with the objective of learning efficient policies able to maximize the collected rewards. The dynamics of the environment are captured by the probability function $p(s_{t+1}, r_{t+1}|s_t, a_t): S \times R \times S \times A \longrightarrow [0, 1]$.

% TODO maybe add that you tried using less RTT info
\subsection{Observation and Action space}

MARLIN utilizes a state representation consisting of 14 features collected during the timeframe following the previous action at step $t-1$. The state space is expanded by computing 7 statistics for each feature, including the last value, mean, standard deviation, minimum, maximum, \gls{ema}, and difference from the previous state $s_{t-1}$. The previous 10 states are stacked together, forming an observation history. Observations are then normalized using a moving average. The action space, instead, comprises continuous actions within the range of $[-1, 1]$, denoting the percentage gain in congestion window (\gls{cwnd}) size. For example, an action of .3 would increase the \gls{cwnd} by 30\%, while -.3 would instead reduce it by the same quantity.

% Deciding how much time passes between the actions is a tough problem
% MARLIN used sRTT-based heuristic
% Time would take too long 
% Fixed time 100ms
\subsection{Timing Decisions} 
    \label{subsec:timing_decisions}
    When designing a \gls{rl}-based \gls{cc} agent, determining \textit{when} the next action is going to be taken constitutes one of the first questions that need to be answered. In the original work \cite{galliera2023marlin}, the authors opted for a heuristic based on the \gls{srtt} detected by the partnering protocol. %allowing the action taken by the agent to affect the channel by several orders of magnitude before taking the next one. 
    This %\gls{srtt}-based 
    approach becomes problematic when the agent is trained within links with delays in the order of seconds, as it would increase the time needed to complete the same training process by several orders of magnitude, without any guarantee of convergence to a policy with the same quality.  

    To address this limitation, our work utilizes a predefined time window of 100ms. Within this timeframe, the partnering protocol collects the necessary statistics, which will be used to form the state of the agent.

% New Reward function penalizing Retransmissions
\subsection{Rewarding the agent in Tactical Networks}
\label{subsec:reward}
    We designed a reward function incorporating penalties based on retransmissions, proving it to be highly suitable to tactical networks, owing to a multitude of compelling reasons. 
    
    By integrating penalties for retransmissions into the reward function, the \gls{rl} agent is actively incentivized to minimize their occurrence, thereby improving the reliability of data transmission. This behavior curtails potential delays and data loss, which harmoniously aligns with the specific requirements and constraints inherent to tactical networks. 
    
    Building upon the principles delineated in \cite{galliera2023marlin} 
    with the controlled training scenarios, it is possible to approximate the traffic influx from competing sources at each time step $t$. Consequently, reward functions can be devised to both confer greater rewards as the agent progressively approaches the optimal utilization of the available bandwidth and greater penalties when retransmissions are present, thereby incentivizing efficient network usage. These concepts can be merged into a single reward function described as follows:
     
    \begin{equation}
    \label{eq:reward_1}
    r_t = -\frac{target_t[1 + r * (1 - loss_c)]}{target_t + acked_t^{cumulative}}
    \end{equation}
    
    \noindent 
    where $target_t$ is an approximation of the number of bytes the agent could have delivered up to step $t$ since the beginning of the episode in order to fully utilize the link and $acked_t^{cumulative}$ represents the number of kilobytes there were acknowledged by the receiver until step $t$. The value $r$ quantifies the number of retransmissions detected by the partnering protocol, while $loss_c$ is the decimal representation of the current random packet loss.

    It is worth mentioning that a strictly negative reward function promotes the agent to accumulate the smallest amount of penalties. The penalty received is smaller the closer the agent gets, at each step, to utilize the link to the best of its possibilities, avoiding overwhelming the other traffic flows.

%% file: sections/experimental_results.tex
\begin{figure*}
\centering    
\includegraphics[width=1\textwidth]{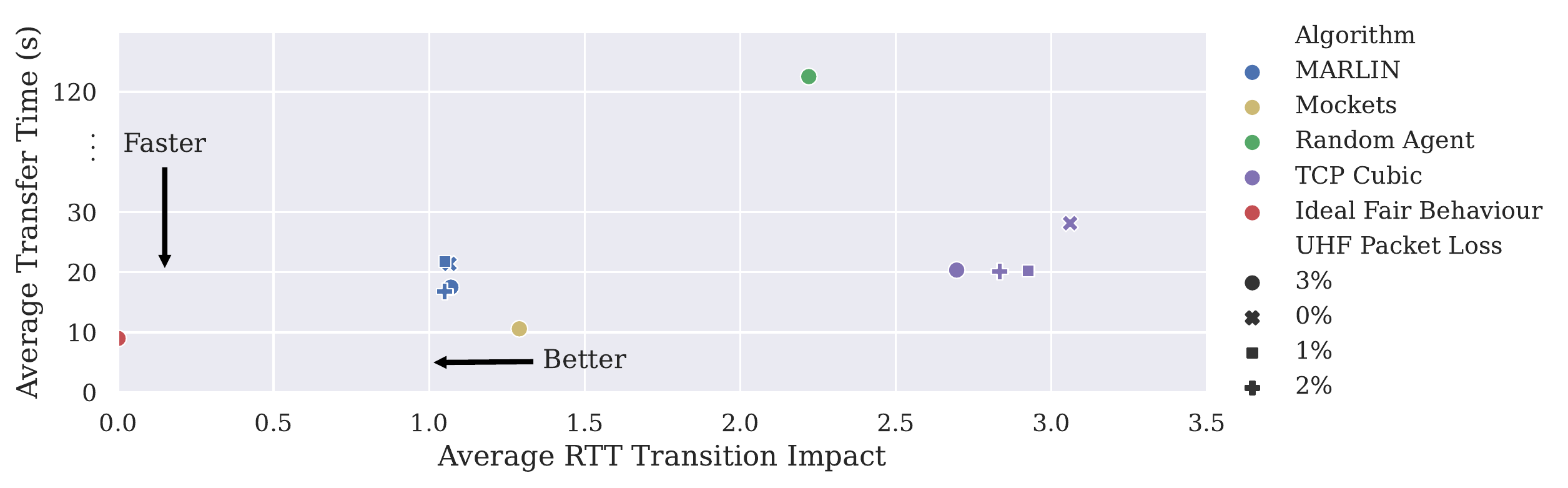}
    \caption{Performance comparison in terms of average transfer time and \gls{rti} across the evaluation episodes.}
    \label{fig:time_comparison}
\end{figure*}

% \subsection{Testbed}
% \label{subsec:testbed}

In order to replicate the wide-ranging variations commonly observed in tactical networks, we have developed a testbed that serves as a training scenario and implements an instance of the tactical environment described in Section \ref{sec:scenario}. 

In this particular scenario, a data rate of 1Mb/s is initially accessible through a \gls{satcom} link, which exhibits a 500ms delay due to the satellite-ground station distance. However, as the tactical network operates in a dynamic environment, it is crucial for it to respond efficiently even in situations where the \gls{satcom} link becomes unavailable or experiences sudden performance degradation. To address such behavior, we simulated a transition to an alternative \gls{uhf} radio link with a 256Kbps data rate, a reduced delay of 125ms, and an assumed 3\% packet loss. Following our framework detailed in Section~\ref{sec:emulated_marlin} it is possible to reproduce such a dynamic environment by defining three elements: the network architecture, the link behavior, and the background traffic patterns.

In Figure~\ref{fig:architecture}, we present the design of a dumbbell architecture that separates two networks, denoted as $LS$ and $RS$. The sender hosts of both Mockets and the \gls{mgen} traffic generator are situated on network $LS$, while their receiving counterparts reside on network $RS$. It is worth noting that the network architecture remains unchanged throughout the scenario.

To replicate the desired scenario, we implement the linking behavior within our network generator module. Initially, the bottleneck link is configured to emulate the properties of the \gls{satcom} link. However, after a duration of 10 seconds, a link transition is triggered, reconfiguring the link to embody the characteristics of the \gls{uhf} radio link. This adjustment accurately reproduces the intended scenario within the testbed.

In the meantime, background traffic flows are emitted by the competing sources, following the sequences presented in Figure~\ref{fig:traffic_patterns}. Each sequence is repeated every 8 seconds and includes two parallel components divided into elephant flows, i.e., long-lived data transfers that represent a large percentage of the total traffic, and mice flows, i.e., short-lived data transfers at low throughput. Elephant flows are alternated every two seconds, while two mice flows are continuously generating very short-lived (in the order of milliseconds) traffic bursts with intervals that follow a Poisson distribution.  

Traffic patterns are adjusted to align with the characteristics of the \gls{uhf} link at the time of transition, as illustrated in Figure~\ref{fig:uhf_traffic}. This effect simulates a perfect adaptation of the background traffic flows to the link change, assuming that competing traffic sources are able to seamlessly adjust to link transitions. It is also supposed that the occurrence of significant \gls{udp} elephant traffic, such as video streams, is interrupted from the competing sources while the communication is routed through the \gls{uhf} link.

\subsection{Training}

When training a \gls{rl} agent, in addition to its objectives, we also need to define the ``horizon'' towards which it is going to optimize its behavior. For this purpose, many \gls{rl} problems define terminal states and the rewards associated with them serve as a signal for successful completion. Once a terminal state is reached, the current episode ends, and a new episode begins. In problems like \gls{cc} it is challenging to define terminal states as, at least theoretically, a communication instated between two nodes could continue indefinitely.
For such a reason, MARLIN considers \gls{cc} a time-unlimited task during training. In these settings, the agent does not optimize its behavior towards reaching a predefined terminal state while maximizing its return, but the focus is shifted to maximizing its returns over an indefinite time horizon. 

Partial episodes, each of which was 200 steps long, were used to diversify the experience of the agent for an approximate duration of 20 seconds per episode, considering that the interval between actions was set to 100 milliseconds, as discussed in Section~\ref{subsec:timing_decisions}.

The algorithms and hyperparameters used mostly followed the default values presented in~\cite{galliera2023marlin}, except for the buffer size and training duration, which were reduced and set respectively to 250K and 500K steps. The observation space and the action space are unaffected compared to the previous work, except for the maximum value that the \gls{cwnd} might assume, here set to 150KB.

\subsection{Evaluation Method}
To evaluate the performance of the trained agent, we conducted a series of file transfers in our experiments. Each evaluation episode involved transmitting a 600KB payload. The experiment, repeated 400 times for both \gls{tcp} and our version of MARLIN, was separated in batches of 100 transfers under various random packet loss probabilities of the \gls{uhf} link, ranging from 0\% to 3\%. It is important to highlight that the agent was exclusively trained using a random packet loss probability of 3\%. The purpose of this evaluation is to confirm that the agent did not learn any patterns based on the number of retransmissions induced by the emulated packet loss.

The results were compared together to the performance of a random agent and Mockets, both performing only on the 3\% packet loss scenario, as they do not adopt any adaptive mechanism. The latter was configured such that the \gls{cwnd} matched the bandwidth capacity of the link, before and after the transition. Enabled by our controlled environment, this seamless process eliminates the necessity of customary protocol reconfiguration, thereby intensifying the challenge for our agent. Finally, we approximated an ideal fair behavior, assuming an algorithm able to split the bandwidth available left by non-collaborative flows (i.e. \gls{udp} traffic) with adaptive flows (i.e. \gls{tcp} flows), without causing any \gls{rtt} increase.

Evaluating \gls{cc} solutions in dynamic and challenging links necessitates careful consideration in the selection of appropriate metrics. In our evaluation process, we focused on three key aspects: the file transfer completion time, the number of retransmissions incurred, and the sensitivity of the agent to link changes and its queue management capabilities.

While the first two factors can be straightforwardly quantified, we believe the latter to be more subtle. Simply relying on the average or maximum \gls{rtt} measured in the link might seem reasonable at first sight, but they would either fail to capture buffer loading peaks or lack information about the specific link where the peak incurred. These observations emphasize the necessity of employing an  able to capture the responsiveness of the agent to adapt to the new link characteristics.

To address this challenge, we designed a novel metric called the \emph{RTT Transition Impact (RTI)}. With the objective of assessing the influence of \gls{cc} algorithms on buffer loading when transitioning between the links, the \gls{rti} leverages the use of the normalized maximum \glspl{rtt} measured separately for each of the $m$ links involved during communication. By considering individual link measurements, the \gls{rti} provides insights into the impact of link changes on the performance of the \gls{cc} algorithm and it can be defined as follows: 

\begin{equation}\label{eq:max_rtt_norm}
    RTI = \ln(\frac{\sum^m_{i=1}\frac{rtt_{i, max}}{rtt_{i, nom}}}{m})
% \begin{split}
%     RTI &= \ln(\frac{\sum^m_{i=1}\eta_{i}}{m}) \\ 
%     \eta_{i} &= \frac{rtt_{i, max}}{rtt_{i, nom}}
% \end{split}
\end{equation}

\noindent
where $m$ identifies the different links involved. The maximum \gls{rtt} detected during the timeframe related to link $i$ is denoted as $rtt_{i, max}$, while the nominal \gls{rtt} for the same timeframe is denoted as $rtt_{i, nom}$. By employing the natural logarithm, we can emphasize the magnitude of the impact for smaller values, while denoting with a value of 0 a flawless performance.

Through the utilization of the \gls{rti} metric, we enhance our evaluation methodology by capturing the intricate dynamics of tactical network environments and their effect on \gls{cc} algorithms. This metric contributes to a more comprehensive understanding of its ability to adapt to evolving link conditions, thereby facilitating a more accurate assessment of \gls{cc} performance in dynamic and challenging scenarios.

\subsection{Results}
Figure~\ref{fig:time_comparison} presents a comparison between MARLIN, Mockets, \gls{tcp} Cubic, a random agent, and the ideal fair behavior in terms of the average transfer time and the average \gls{rti} achieved. The results show that our trained agent exhibits a moderate improvement in file transfer completion time compared to \gls{tcp}, with an average performance of 19.3s compared to 22.20s. However, it is noticeably slower than Mockets, which achieves an average completion time of 10.59s. This outcome is expected as Mockets is purposefully designed for tactical environments and adopts an aggressive policy that introduces high overhead on the network, as we will further discuss.

In terms of average \gls{rti}, MARLIN consistently outperforms its competitors across different scenarios, regardless of varying packet losses and with minimal \gls{rti} fluctuations between the 4 scenarios. On the other hand, \gls{tcp} Cubic exhibits a peculiar and variable behavior in terms of \gls{rti} performance. It is observed that lower levels of packet loss result in higher \gls{rti} for \gls{tcp} Cubic, indicating potential limitations in its queue management capabilities. This contrast highlights the robustness and stability of MARLIN in maintaining low \gls{rti} values in our scenario, showcasing promising performance for tactical network environments.

\begin{figure}[t]
    \centering
    \includegraphics[width=0.75\columnwidth]{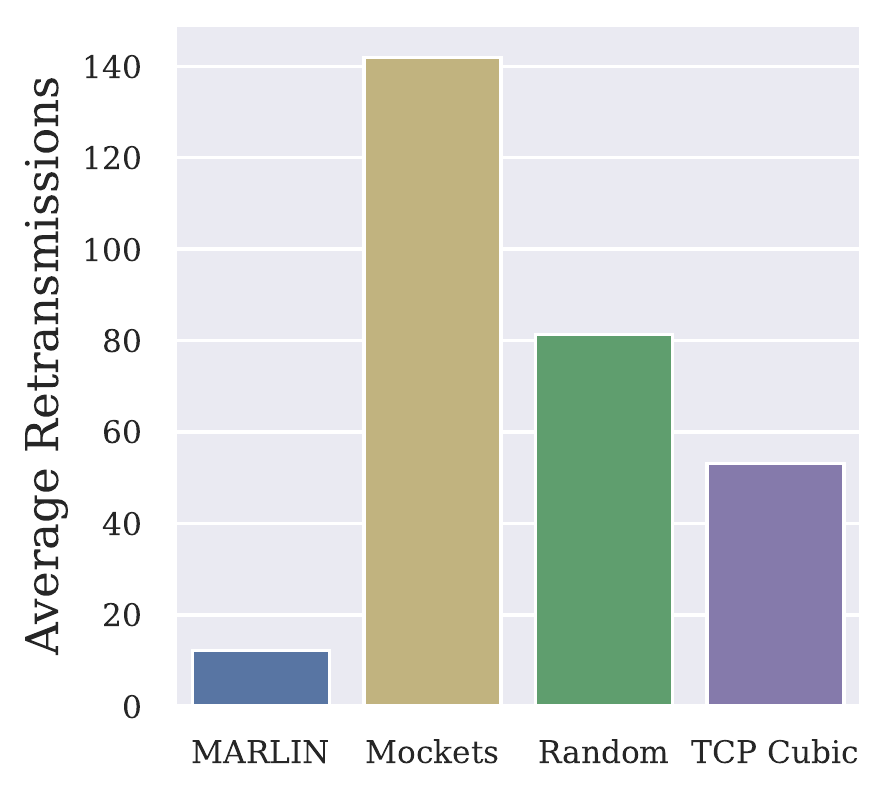}
    \caption{The average number of retransmissions during evaluation with a 3\% random packet loss set on the UHF link.}
    \label{fig:retransmissions}
\end{figure}

To conclude and support our results, we assessed the average number of retransmissions per episode involving MARLIN, Mockets, the random agent, and \gls{tcp} Cubic when a 3\%  packet loss was present on the \gls{uhf} link. Figure~\ref{fig:retransmissions} illustrates how our agent is able to limit the average retransmissions to 12.22 against the 53.1 for \gls{tcp} Cubic and 141.93 for Mockets.

%% file: sections/related_work.tex
Considerable emphasis has consistently been placed on the optimization of \gls{cc} algorithms. This has led to the proposal of numerous heuristics, including notable examples such as \gls{tcp} Vegas \cite{Brakmo1994TCPVN}, BBR~\cite{bbr}, and \gls{tcp} CUBIC~\cite{cubic}. However, most algorithms achieve satisfactory performance in specific scenarios often demonstrating poor generalization, especially in dynamic and unreliable networks \cite{chaudhary2017}.

In the ever-evolving field of networking, it is not surprising that recent efforts have focused also on \gls{rl}-based optimization techniques, aiming to develop solutions able to adapt to the complexity and heterogeneity of networking environments.
Approaches such as DRL-CC~\cite{drl-cc} have focused on training  policies in simulated environments, achieving competitive performance, but demonstrating several challenges when applied to real networks. Others, such as MVFST-RL~\cite{Sivakumar2019} focused on training non-blocking \gls{cc} \gls{rl} agents on the application level, overcoming the limitation of non-blocking solutions. MARLIN~\cite{galliera2023marlin}, followed a similar non-blocking approach, with a particular emphasis on training and evaluating the agent end-to-end on a real network.  

To the best of our knowledge, 
this paper presents a two-fold novel contribution: it is the first \gls{rl} environment for \gls{cc} with a focus on tactical networks and the first exploiting the flexibility of emulated networks with containerized applications to train and evaluate a \gls{rl} agent for \gls{cc}.

%% file: sections/conclusion.tex
%Tactical networks are highly dynamic networking environments calling for network communication protocols capable of adapting to ever-changing conditions such as transmission latency and bandwidth availability. 

This paper presented a step toward comprehensive \gls{rl} environments capable of training policies on emulated networks with dynamic link behavior, paving the way for studying the application of \gls{rl} to complex networking scenarios, which might be impractical or costly in real network environments.

%Among the other criteria used for evaluation, we have also introduced the \gls{rti}, a metric useful for capturing the sensitivity of the agent to link changes and its queue management capabilities. 

%Finally, our 
Experimental results demonstrated how MARLIN, with a relatively exiguous training budget, was able to learn an effective \gls{cc} policy in the presented scenario, addressing the shortcomings of both the Mockets' aggressive strategy and \gls{tcp} Cubic. Among the other criteria used for evaluation, we have also introduced the \gls{rti}, a metric useful for capturing the sensitivity of the agent to link changes. 

It is crucial to mention that the framework and the results presented here, albeit we believe them to be promising and compelling, are still limited and focus on a specific training scenario. A variety of approaches and methods still need to be explored, implemented, and evaluated within MARLIN, which will be part of our future work. These include but are not limited to, parallel environments, more heterogeneous dynamic network setups, self-timed decisions, and competitive and/or cooperative \gls{marl} settings.

%The path toward a comprehensive, competitive, and generalizable \gls{rl} agent for \gls{cc} is still long and rich in exciting challenges and opportunities. 

%% file: main.bbl
\begin{thebibliography}{10}
\providecommand{\url}[1]{#1}
\csname url@samestyle\endcsname
\providecommand{\newblock}{\relax}
\providecommand{\bibinfo}[2]{#2}
\providecommand{\BIBentrySTDinterwordspacing}{\spaceskip=0pt\relax}
\providecommand{\BIBentryALTinterwordstretchfactor}{4}
\providecommand{\BIBentryALTinterwordspacing}{\spaceskip=\fontdimen2\font plus
\BIBentryALTinterwordstretchfactor\fontdimen3\font minus
  \fontdimen4\font\relax}
\providecommand{\BIBforeignlanguage}[2]{{%
\expandafter\ifx\csname l@#1\endcsname\relax
\typeout{** WARNING: IEEEtran.bst: No hyphenation pattern has been}%
\typeout{** loaded for the language `#1'. Using the pattern for}%
\typeout{** the default language instead.}%
\else
\language=\csname l@#1\endcsname
\fi
#2}}
\providecommand{\BIBdecl}{\relax}
\BIBdecl

\bibitem{Sivakumar2019}
\BIBentryALTinterwordspacing
V.~Sivakumar, O.~Delalleau, T.~Rocktäschel, A.~H. Miller, H.~Küttler,
  N.~Nardelli, M.~Rabbat, J.~Pineau, and S.~Riedel, ``Mvfst-rl: An asynchronous
  rl framework for congestion control with delayed actions,'' 10 2019.
  [Online]. Available: \url{http://arxiv.org/abs/1910.04054}
\BIBentrySTDinterwordspacing

\bibitem{drl-cc}
Z.~Xu, J.~Tang, C.~Yin, Y.~Wang, and G.~Xue, ``Experience-driven congestion
  control: When multi-path tcp meets deep reinforcement learning,'' \emph{IEEE
  Journal on Selected Areas in Communications}, vol.~37, no.~6, pp. 1325--1336,
  2019.

\bibitem{galliera2023marlin}
R.~Galliera, A.~Morelli, R.~Fronteddu, and N.~Suri, ``Marlin: Soft actor-critic
  based reinforcement learning for congestion control in real networks,'' in
  \emph{NOMS 2023-2023 IEEE/IFIP Network Operations and Management Symposium},
  2023, pp. 1--10.

\bibitem{cubic}
\BIBentryALTinterwordspacing
S.~Ha, I.~Rhee, and L.~Xu, ``Cubic: A new tcp-friendly high-speed tcp
  variant,'' \emph{SIGOPS Oper. Syst. Rev.}, vol.~42, no.~5, p. 64–74, jul
  2008. [Online]. Available: \url{https://doi.org/10.1145/1400097.1400105}
\BIBentrySTDinterwordspacing

\bibitem{mockets1}
E.~Benvegnù, N.~Suri, M.~Tortonesi, and T.~Esterrich, ``Seamless network
  migration using the mockets communications middleware,'' in \emph{2010 -
  MILCOM 2010 MILITARY COMMUNICATIONS CONFERENCE}, 2010, pp. 2298--2303.

\bibitem{mockets2}
A.~Morelli, M.~Provosty, R.~Fronteddu, and N.~Suri, ``Performance evaluation of
  transport protocols in tactical network environments,'' in \emph{MILCOM 2019
  - 2019 IEEE Military Communications Conference (MILCOM)}, 2019, pp. 30--36.

\bibitem{anglova}
N.~Suri, A.~Hansson, J.~Nilsson, P.~Lubkowski, K.~Marcus, M.~Hauge, K.~Lee,
  B.~Buchin, L.~Mısırhoğlu, and M.~Peuhkuri, ``A realistic military scenario
  and emulation environment for experimenting with tactical communications and
  heterogeneous networks,'' in \emph{2016 International Conference on Military
  Communications and Information Systems (ICMCIS)}, 2016, pp. 1--8.

\bibitem{peuster2016medicine}
M.~Peuster, H.~Karl, and S.~van Rossem, ``Medicine: Rapid prototyping of
  production-ready network services in multi-pop environments,'' in \emph{2016
  IEEE Conference on Network Function Virtualization and Software Defined
  Networks (NFV-SDN)}, Nov 2016, pp. 148--153.

\bibitem{mgen}
N.~R. L. N. P. E. A. N. P.~R. Group, ``Multi-generator (mgen) network test
  tool,''
  \url{https://www.nrl.navy.mil/Our-Work/Areas-of-Research/Information-Technology/NCS/MGEN/},
  2021.

\bibitem{suttonBarto}
R.~S. Sutton and A.~G. Barto, \emph{Reinforcement Learning: An
  Introduction}.\hskip 1em plus 0.5em minus 0.4em\relax Cambridge, MA, USA: A
  Bradford Book, 2018.

\bibitem{Brakmo1994TCPVN}
L.~S. Brakmo, S.~W. O'Malley, and L.~L. Peterson, ``Tcp vegas: New techniques
  for congestion detection and avoidance,'' in \emph{SIGCOMM}, 1994.

\bibitem{bbr}
\BIBentryALTinterwordspacing
N.~Cardwell, Y.~Cheng, C.~S. Gunn, S.~H. Yeganeh, and V.~Jacobson, ``Bbr:
  Congestion-based congestion control,'' \emph{ACM Queue}, vol. 14,
  September-October, pp. 20 -- 53, 2016. [Online]. Available:
  \url{http://queue.acm.org/detail.cfm?id=3022184}
\BIBentrySTDinterwordspacing

\bibitem{chaudhary2017}
P.~Chaudhary and S.~Kumar, ``Comparative study of tcp variants for congestion
  control in wireless network,'' in \emph{2017 International Conference on
  Computing, Communication and Automation (ICCCA)}, 2017, pp. 641--646.

\end{thebibliography}
